%% file: main.tex
\documentclass[10pt,twocolumn,letterpaper]{article}

\usepackage[pagenumbers]{cvpr} 

\input{preamble}

\definecolor{cvprblue}{rgb}{0.21,0.49,0.74}
\usepackage[pagebackref,breaklinks,colorlinks,allcolors=cvprblue]{hyperref}

\def\paperID{*****} 
\def\confName{CVPR}
\def\confYear{2026}

\title{Beyond Motion Primitives: Behavioral Activity Recognition from Head-Mounted IMU}

\author{
\parbox{0.97\textwidth}{\centering
Chung-Ta~Huang\textsuperscript{*}, L\'{e}opold~Das\textsuperscript{*}, Jeffrey~Zhou,
Faizaan~Siddique\\[-0.05em]
Julia~Seungjoo~Baek, Serena~Liu, Andrew~Rusli, Todd~Y.~Zhou\\[-0.05em]
Freddy~Yu, Sinclair~Hansen, Ziling~Hu, Arnav~Sharma, and
Mengyu~Wang\textsuperscript{\textdagger}\\[0.45em]
Harvard AI and Robotics Lab, Harvard University\\[0.25em]
{\small \href{mailto:chungta_huang@gsd.harvard.edu}{\nolinkurl{chungta_huang@gsd.harvard.edu}} \quad
\textsuperscript{\textdagger}Corresponding author:~
\href{mailto:mengyu_wang@meei.harvard.edu}{\nolinkurl{mengyu_wang@meei.harvard.edu}}}}
}

\begin{document}
\maketitle
\input{sec/0_abstract}
\input{sec/1_intro}
\input{sec/2_related}
\input{sec/3_dataset}
\input{sec/4_method}
\input{sec/5_experiments}
\input{sec/6_limits}
\input{sec/8_conclusion}

{
    \small
    \bibliographystyle{ieeenat_fullname}
    \bibliography{main}
}

\input{sec/X_suppl}

\end{document}

%% file: preamble.tex
\usepackage{booktabs}
\usepackage{multirow}

\usepackage{amsmath}
\usepackage{amssymb}

\usepackage{graphicx}
\usepackage{subcaption}

\newcommand{\TODO}[1]{\textbf{\color{red}[TODO: #1]}}
\renewcommand{\TODO}[1]{}

\usepackage{xspace}

\usepackage{enumitem}
\setlist{nosep,leftmargin=*}

%% file: sec/0_abstract.tex
\begin{abstract}

AR smart glasses need continuous behavioral context to offer proactive assistance, yet their most practical always-on sensor, the head-mounted Inertial Measurement Unit (IMU), detects only motion primitives such as walking or standing.
We push beyond motion primitives to behavioral-level recognition, defining five categories that balance AR application need with sensor observability.
To this end, we construct a 160K-sample Ego4D dataset with a four-tier quality assurance framework spanning 8 activity scenarios, and propose HiT-HAR, a 703K-parameter hierarchical model that outperforms prior head-mounted IMU architectures including IMU2CLIP and a fine-tuned Mantis foundation model on five-class action and eight-class scenario recognition.
We further map the observability frontier of head-mounted IMU through per-class separability analysis, identifying which behavioral categories are reliably observable (Locomotion), which benefit from temporal context (Object Transfer, Task Operation), and where scenario-dependent signal overlap poses remaining challenges.
Our results indicate that architectural choices exploiting temporal context and scenario structure outperform simply scaling model size.

\end{abstract}

%% file: sec/1_intro.tex
\section{Introduction}
\label{sec:intro}

AR assistance that goes beyond passive display requires the system to understand the user's behavioral context: not just whether they are moving, but what they are functionally doing.
Head-mounted inertial measurement units (IMUs) offer a privacy-preserving, always-on window into user behavior, and as AR glasses move toward consumer deployment, the embedded IMU becomes the most practical sensor for continuously sensing behavioral context without the battery cost of always-on cameras.

Yet most IMU-based activity recognition targets high-momentum motion primitives such as walking, running, and sitting~\cite{egocharm2025,imu2clip2023,comodo2025}.
These categories are well-separated in accelerometer space but tell an AR assistant little about \emph{what} the user is functionally doing.
Knowing that a user is walking says nothing about whether they are searching for a tool or transferring materials; an AR system needs exactly this behavioral distinction to decide what assistance to offer.

Consider a user performing a mechanical repair.
Over 30 seconds, they pick up a wrench (Object Transfer), tighten a bolt (Task Operation), pause to inspect their work (Stationary), then walk to get more parts (Locomotion).
An AR assistant that only detects ``walking'' vs.\ ``stationary'' cannot distinguish these functional states.

We define five behavioral categories that balance what AR applications need against what the IMU can observe, then systematically map the observability boundaries of head-mounted IMU using Ego4D~\cite{ego4d2022} data.


Our contributions are threefold.
\begin{enumerate}[leftmargin=1.5em,itemsep=2pt]
    \item \textbf{Annotated dataset with quality framework.}
    A 160K-sample annotated dataset from Ego4D head-mounted IMU spanning 8 activity scenarios, with 27K gold labels from 12 annotators and a four-tier quality framework (Sec.~\ref{sec:dataset}).

    \item \textbf{HiT-HAR (Hierarchical Temporal Human Activity Recognition).}
    A 703K-parameter architecture combining multi-dilation CNN-GRU local encoding with Transformer temporal aggregation and a scenario-informed gated action head, outperforming prior head-mounted IMU architectures including IMU2CLIP~\cite{imu2clip2023} on five-class action F1 (Table~\ref{tab:main}).

    \item \textbf{Observability frontier analysis.}
    A systematic mapping of which behavioral categories are reliably observable from head-mounted IMU, which benefit from temporal context, and where scenario-dependent signal overlap poses remaining challenges (Sec.~\ref{sec:observability}).
\end{enumerate}

%% file: sec/2_related.tex
\section{Related Work}
\label{sec:related}
\begin{figure*}[!t]
    \centering
    \includegraphics[width=0.8\linewidth,height=0.3\textheight]{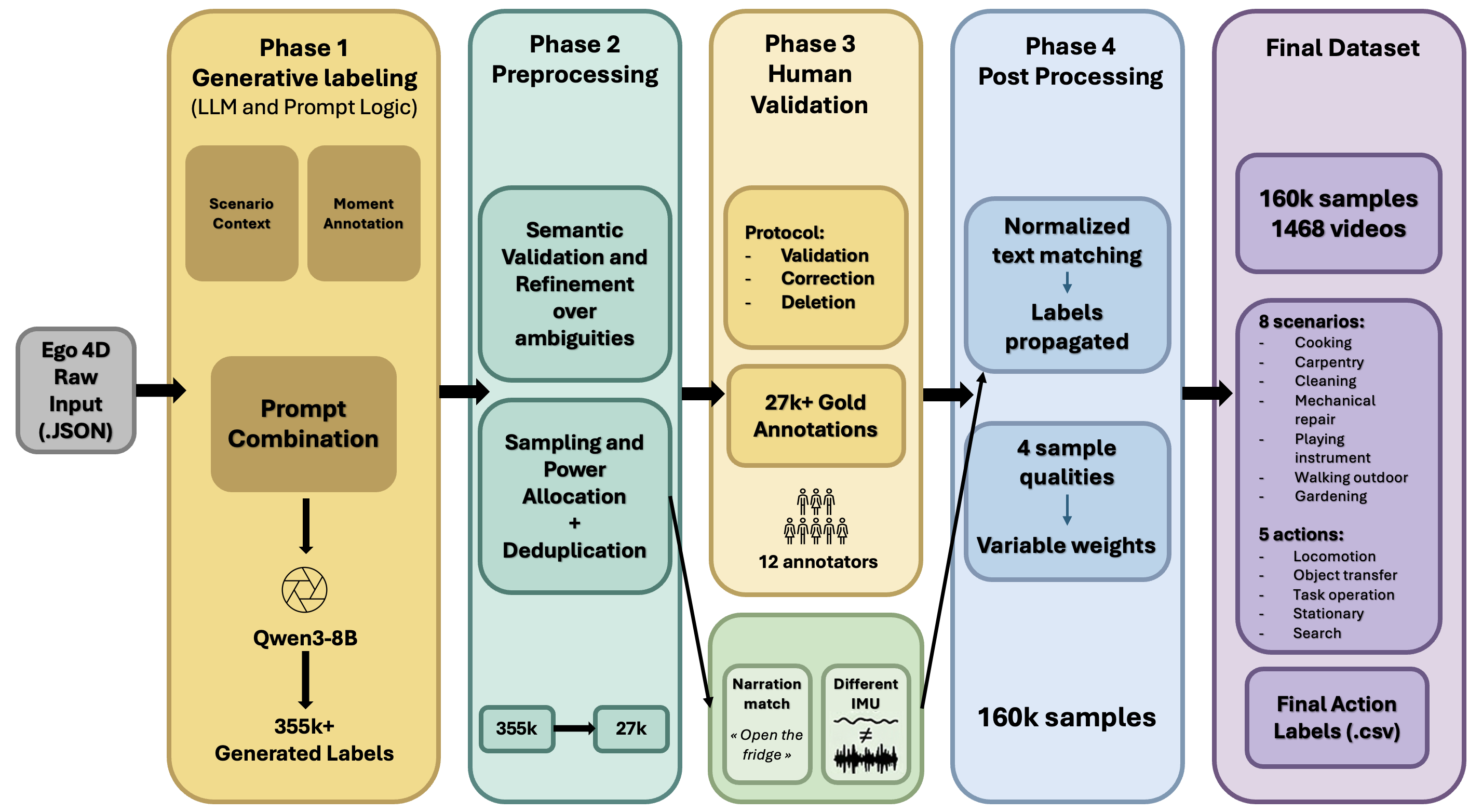}
    \caption{End-to-end data pipeline: LLM-generated labels are preprocessed, verified by 12 human annotators producing 27K gold annotations, then propagated to near-duplicate narrations yielding ${\sim}$160K labeled samples with four quality tiers.}
    \label{fig:data_diagram}
\end{figure*}
\paragraph{Activity taxonomy and the behavioral level.}
HAR research organizes movements into hierarchical granularity levels~\cite{moeslund2006survey,bulling2014tutorial,morshed2023taxonomy}.
Motion primitives (walking, standing) are well-studied and reliably detectable from inertial sensors, while fine-grained manipulation (hammering, pouring) typically requires vision or multi-sensor fusion~\cite{pareek2021survey}.
Multi-level annotation schemes such as OPPORTUNITY++~\cite{opportunitypp2022} formalize this hierarchy for sensor datasets, defining postures, gestures, and high-level activities as separate label layers.
Between these extremes lies a \emph{behavioral level} that captures functional intent: not how the user moves, but what they are trying to accomplish.
However, prior taxonomies are designed primarily around signal separability or annotation convenience rather than downstream application need.
Designing taxonomies at this level requires balancing application relevance against sensor observability, a trade-off we formalize through dual-criteria taxonomy design (Sec.~\ref{sec:dataset:taxonomy}).

\paragraph{Head-mounted IMU recognition.}
EgoCHARM~\cite{egocharm2025} introduces a hierarchical architecture pairing a per-window encoder with a sequence-level aggregator, classifying 3 motion primitives and 9 activity scenarios from a single egocentric IMU (${\sim}$85K params).
IMU2CLIP~\cite{imu2clip2023} aligns IMU embeddings with CLIP for zero-shot classification and has become a standard baseline in head-mounted IMU recognition, adopted by EgoCHARM, PRIMUS~\cite{primus2025}, and COMODO~\cite{comodo2025}.
COMODO distills video supervision into an IMU encoder on Ego4D~\cite{ego4d2022}.
MopFormer~\cite{mopformer2025} applies a Transformer encoder to wearable-sensor motion primitives, showing that self-attention over temporal windows outperforms purely recurrent models, though it targets body-worn sensors and coarse motion categories rather than head-mounted behavioral recognition.
Haresamudram~\etal~\cite{haresamudram2025limits} showed that language supervision underperforms standard training for sensor HAR, motivating task-specific architectural choices over pre-training at scale.
However, none of these targets the behavioral level — the functional intent behind motion patterns.

\paragraph{Architectural techniques for IMU HAR.}
Per-channel recalibration via Squeeze-and-Excitation (SE) attention~\cite{hu2018squeeze} and multi-dilation convolutions~\cite{mopformer2025} capture motion patterns at multiple time scales; our Window-Level Encoder adopts both.
Gated Multimodal Networks~\cite{arevalo2020gated} learn multiplicative gates to blend signals from different sources.
We repurpose gated fusion for blending local per-window and contextual sequence-level representations within a single-sensor pipeline, guided by the observation that class separability varies by scenario.

\paragraph{Datasets and annotation for IMU HAR.}
Existing head-mounted IMU benchmarks derive labels from Ego4D scenario metadata~\cite{ego4d2022} or motion-primitive ontologies~\cite{egocharm2025,opportunitypp2022}, but none provide fine-grained behavioral action labels from egocentric IMU.
IMUGPT~\cite{leng2024imugpt} demonstrated that LLMs can generate plausible action labels at scale, yet without human verification the resulting labels are noisy and inconsistent.
We build on this paradigm with an LLM-human backfeed loop: Qwen3-8B generates initial labels over 355K narrations, then human annotators produce 27K gold annotations with a four-tier quality framework (Sec.~\ref{sec:dataset:annotation}).

%% file: sec/3_dataset.tex
\section{Dataset and Quality Framework}
\label{sec:dataset}

We construct an annotated dataset from the Ego4D egocentric video corpus~\cite{ego4d2022}, pairing head-mounted 6-axis IMU recordings (accelerometer and gyroscope, 50\,Hz) with behavioral-level action labels.
The dataset spans 8 activity scenarios across 1{,}468 videos, totaling ${\sim}$160K labeled samples with a four-tier quality framework.

\begin{figure}[b]
    \centering
    \includegraphics[width=\linewidth]{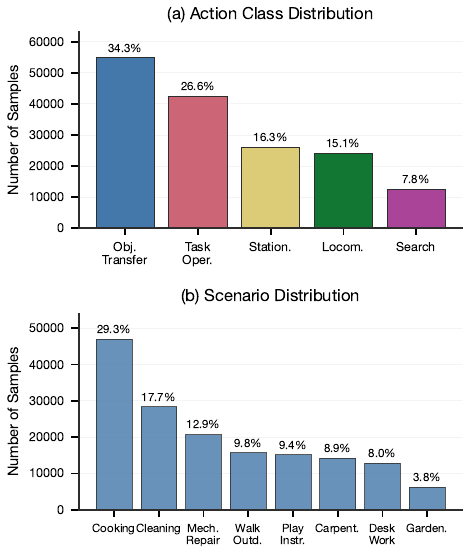}
    \caption{Dataset distribution over the full 160K-sample dataset. (a) Five action classes. (b) Eight Ego4D activity scenarios.}
    \label{fig:distribution}
\end{figure}


\subsection{Five-Class Behavioral Taxonomy}
\label{sec:dataset:taxonomy}

Following the HAR hierarchy~\cite{moeslund2006survey,bulling2014tutorial}, we target a \emph{behavioral level} between motion primitives (walking, standing) and fine-grained manipulation (hammering, pouring).
The taxonomy balances two criteria: \emph{application relevance}, where each class corresponds to a distinct AR assistance action enabling a concrete system response, and \emph{observability hypothesis}, where each class has a plausible head-motion signature distinguishing it from others in at least some activity scenarios.

We derive five categories by clustering Ego4D narration verbs according to semantic similarity, filtering for application relevance, and assessing the observability hypothesis against head-motion intuition (Table~\ref{tab:classes}).
Object Transfer (picking up, carrying) triggers spatial guidance and produces intermittent head turns toward targets.
Task Operation (tightening, cutting) triggers step-by-step prompts and involves a steadier, task-focused gaze.
Stationary (standing idle, observing) signals an opportunity for ambient information display and produces low energy across all channels.
Locomotion (walking, climbing stairs) triggers navigation assistance and is characterized by periodic gait acceleration.
Search (scanning a shelf, looking around) corresponds to a common trigger for proactive AR assistance.
We hypothesize that it involves distinctive head rotation, yet some instances instead exhibit static gaze, producing signals with limited head movement.
We include Search despite its limited observability because of its high application value and validate this observability gap in Sec.~\ref{sec:observability}.

\begin{figure}[!t]
    \centering
    \includegraphics[width=\linewidth]{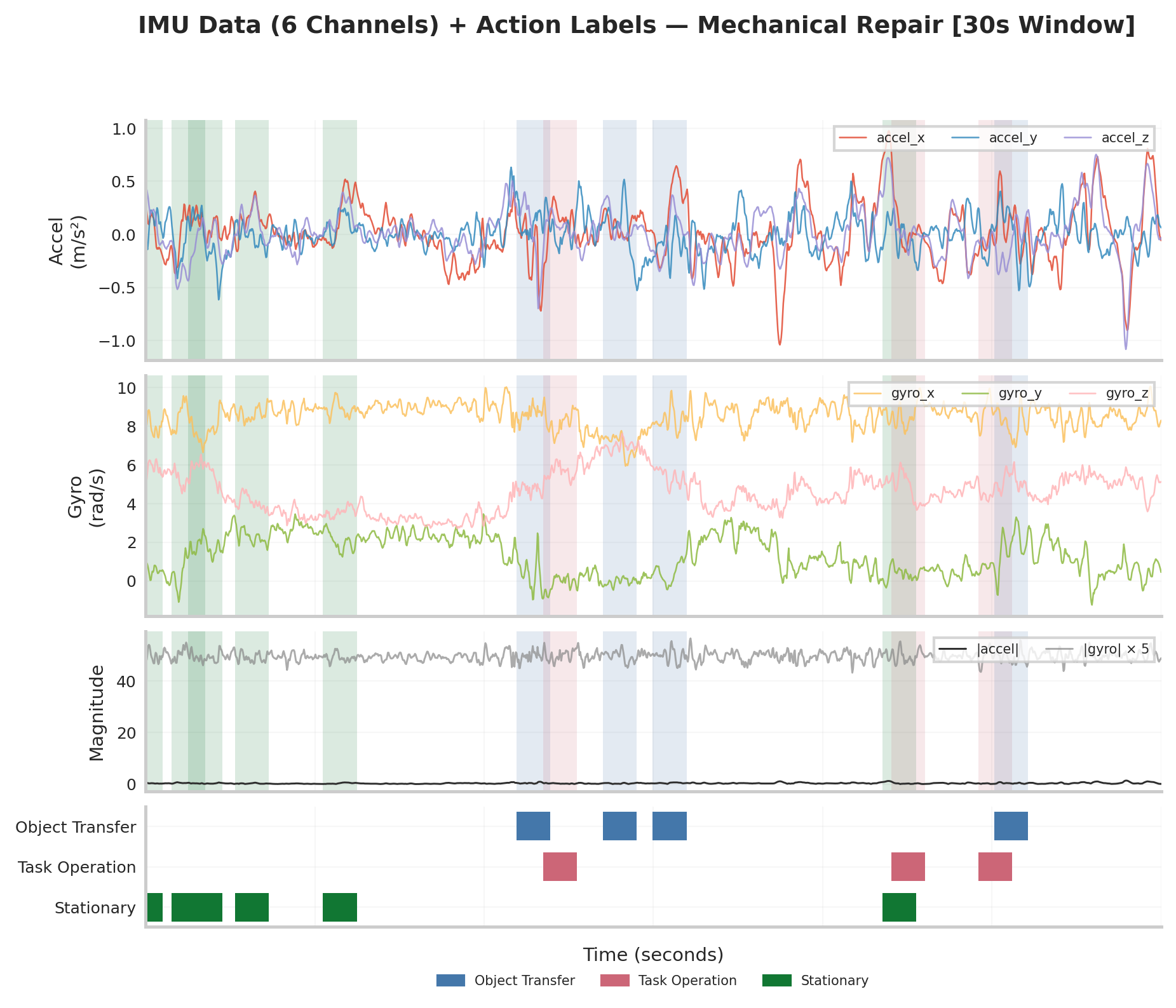}
    \caption{Head-mounted IMU signals (30-second window, Mechanical Repair) with 5-class action labels. Frequent transitions between Object Transfer, Task Operation, and Stationary illustrate behavioral diversity within a single scenario.}
    \label{fig:imu_example}
\end{figure}

The dataset spans 8 Ego4D scenarios (Cooking, Carpentry, Cleaning, Desk Work, Mechanical Repair, Playing Instrument, Walking Indoors, Walking Outdoors).
Fig.~\ref{fig:imu_example} illustrates how these behavioral classes map to head-mounted IMU signals in a representative scenario.

\begin{table}[t]
\centering
\caption{Five-class behavioral taxonomy with per-class distribution (full 160K-sample dataset).}
\label{tab:classes}
\small
\setlength{\tabcolsep}{4pt}
\begin{tabular}{@{}lp{3.8cm}rr@{}}
\toprule
\textbf{Class} & \textbf{Example narration} & \textbf{Count} & \textbf{\%} \\
\midrule
Object Transfer & ``picks up the wrench'' & 54{,}897 & 34.3 \\
Task Operation  & ``tightens the bolt'' & 42{,}499 & 26.6 \\
Stationary      & ``stands idle'' & 26{,}012 & 16.3 \\
Locomotion      & ``walks to the shelf'' & 24{,}104 & 15.1 \\
Search          & ``scans the shelf for a pan'' & 12{,}487 & 7.8 \\
\bottomrule
\end{tabular}
\end{table}

\subsection{LLM-Human Backfeed Annotation}
\label{sec:dataset:annotation}

Our annotation follows an \emph{LLM-human backfeed loop}~\cite{wang2024lapras}: Qwen3-8B, an 8-billion-parameter reasoning model, classifies 355K Ego4D narrations into five categories given the narration text, activity scenario, and taxonomy definitions, producing per-sample reasoning chains.
Following the LLM-assisted paradigm also explored in IMUGPT~\cite{leng2024imugpt}, 12 human annotators then verify labels across two rounds using the narration, scenario context, and LLM reasoning as evaluation inputs, producing 27{,}355 gold annotations.

\paragraph{Sampling strategy:}
To oversample rare classes relative to proportional sampling, we use square-root frequency resampling ($\alpha{=}0.5$), which increases the annotation budget for minority classes such as Search.
Narration strings repeat across the corpus: many distinct (video, time) rows share the same text.
To avoid redundant annotation, we deduplicate narrations that share both normalized text (lowercased, articles and trailing punctuation removed) and LLM-assigned label; narrations with the same text but different labels remain separate.

\paragraph{Verified label propagation:} Verified labels are propagated to near duplicates (for their IMU diversity), to yield 160K total samples across 1{,}468 videos (5.8$\times$ expansion).
The LLM achieves 92.7\% agreement with gold labels.

\paragraph{Key finding:} Of 470 multi-label conflicts, 85.9\% stem from taxonomy boundary ambiguity rather than LLM errors.
The dominant conflict pairs involve Object Transfer/Task Operation and Search/Stationary, consistent with the observability hypothesis that these class pairs share similar head-motion signatures in certain scenarios.

\subsection{Quality Tier System}
\label{sec:dataset:quality}

Not all gold labels are equally reliable.
Following work on learning under temporal label noise~\cite{nagaraj2025tenor}, we assign four quality tiers based on annotator behavior signals: Tier 1 (high confidence, 30.9\%) requires a Gold verdict with no secondary choice and an unambiguous verb (weight 1.0); Tier 2 (moderate, 35.0\%) allows a secondary choice or ambiguous verb (weight 0.8); Tier 3 (corrected, 9.6\%) captures cases where annotators corrected the LLM label (weight 0.5--0.7); and Tier 4 (excluded, 24.5\%) covers skipped or deleted samples (weight 0.0).
These confidence weights modulate the focal loss~\cite{lin2017focal} during training (Sec.~\ref{sec:method:training}).

\paragraph{Label sparsity:}
Action labels cover 17.4\% of total video time after propagation, reflecting the inherent sparsity of narration-based annotation in Ego4D.
This sparsity, combined with the narration-to-IMU semantic gap (visual narrations describe hand-level events like ``picks up the cup'' while the IMU captures only head motion), motivates a model that exploits temporal context and multi-task scenario supervision.

%% file: sec/4_method.tex
\section{Method}
\label{sec:method}

The challenges above — overlapping class distributions in raw IMU space, sparse labels, and the narration-to-IMU semantic gap — motivate three design choices: multi-dilation convolutions to capture motion at multiple time scales, a Transformer-based sequence aggregator to exploit 30 seconds of context for resolving single-window ambiguity, and scenario-informed gating to leverage scenario-dependent separability.

\subsection{Architecture Overview}
\label{sec:method:overview}

\begin{figure*}[t]
    \centering
    \includegraphics[width=0.65\textwidth]{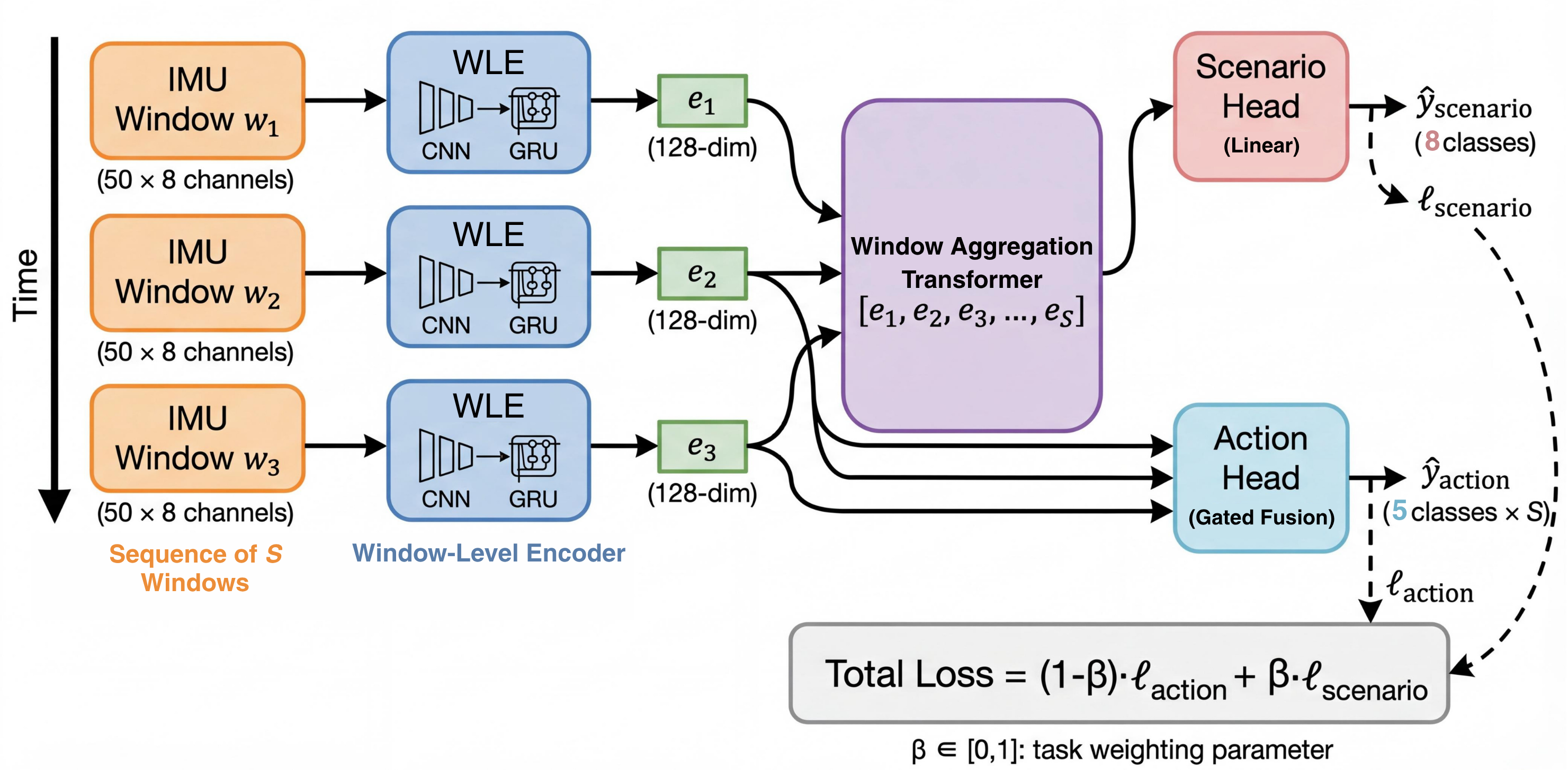}
    \caption{HiT-HAR architecture. The Window-Level Encoder (WLE) processes each 1-second IMU window independently via CNN and GRU, producing 128-dimensional embeddings. The Window Aggregation Transformer (WAT) aggregates a sequence of window embeddings. A scenario head predicts from the CLS token, while a gated action head fuses local and contextual predictions. Total: 703K parameters.}
    \label{fig:architecture}
\end{figure*}

HiT-HAR is a lightweight hierarchical model for joint scenario classification and behavioral-level action recognition from head-mounted IMU (Fig.~\ref{fig:architecture}).
Following the hierarchical encoder-aggregator design introduced by EgoCHARM~\cite{egocharm2025}, a \emph{Window-Level Encoder} (WLE) maps each 1-second IMU window to a 128-dimensional embedding.
A \emph{Window Aggregation Transformer} (WAT) aggregates 30 such embeddings, capturing 30 seconds of temporal context.
Two prediction heads operate on the aggregated representation: a scenario head and a gated action head.

\subsection{Window-Level Encoder}
\label{sec:method:wle}

The WLE processes each 1-second window of 8-channel input (6-axis IMU plus acceleration and gyroscope norms) at 50\,Hz.
A stem convolution projects the input to 64 channels, followed by three multi-dilation CNN blocks with dilations $\{1,2,4\}$ and channel recalibration via Squeeze-and-Excitation attention~\cite{hu2018squeeze}.
A bidirectional GRU with attention pooling yields the 128-dimensional per-window embedding $\mathbf{e}_t$.

\subsection{Window Aggregation Transformer}
\label{sec:method:wat}

The WAT aggregates $S{=}30$ window embeddings into sequence-level and per-window contextual representations using a single-layer Transformer encoder with a learnable CLS token, learnable positional embeddings, 4 attention heads, and feed-forward dimension 512.
The CLS output $\mathbf{h}_{\text{cls}}$ serves as the sequence-level embedding; per-window outputs $\mathbf{h}_t$ provide contextual representations spanning the full 30-second window.
The 30-second context spans approximately six typical action transitions in Ego4D narrations.

\subsection{Gated Action Head}
\label{sec:method:head}

Two heads produce predictions from the WAT outputs.
The \emph{scenario head} applies a linear classifier to $\mathbf{h}_{\text{cls}}$, predicting one of eight scenarios.
The \emph{gated action head} fuses local (WLE) and contextual (WAT) signals for per-window action prediction:
\begin{align}
    \mathbf{a}_{\text{loc}} &= W_{\text{loc}}\,\mathbf{e}_t, \quad
    \mathbf{a}_{\text{ctx}} = W_{\text{ctx}}\,\mathbf{h}_t, \label{eq:local_ctx} \\
    g &= \sigma\!\bigl(\text{MLP}([\mathbf{e}_t;\, \mathbf{h}_t])\bigr), \label{eq:gate} \\
    \mathbf{a}_t &= (1 - g)\,\mathbf{a}_{\text{loc}} + g\,\mathbf{a}_{\text{ctx}}, \label{eq:fusion}
\end{align}
where $g \in [0,1]$ is a learned scalar gate that adaptively blends local motion evidence with longer-range context.
The scenario classification head supervises $\mathbf{h}_{\text{cls}}$, which participates in self-attention with all $\mathbf{h}_t$ tokens, encouraging $\mathbf{h}_t$ to encode scenario-discriminative structure that the gate can exploit.

\subsection{Multi-Task Training}
\label{sec:method:training}

The total loss combines scenario and action objectives:
\begin{equation}
    \mathcal{L} = \beta \cdot \mathcal{L}_{\text{scenario}} + (1 - \beta) \cdot \mathcal{L}_{\text{action}}
    \label{eq:loss}
\end{equation}
where $\beta = 0.3$ balances scenario and action objectives (Sec.~\ref{sec:experiments:ablation}).
Both losses use focal loss~\cite{lin2017focal} ($\gamma{=}2.0$) with inverse-frequency class weights.
We train with AdamW, cosine scheduling, gradient clipping (max norm 1.0), EMA (decay 0.999), and label smoothing ($\epsilon{=}0.05$).

%% file: sec/5_experiments.tex
\section{Experiments}
\label{sec:experiments}

\subsection{Experimental Setup}
\label{sec:experiments:setup}

We train and evaluate on our Ego4D-derived dataset (Sec.~\ref{sec:dataset}), splitting 1{,}468 videos by UID into train (111K samples), validation (24K), and test (25K) partitions.
Our primary metric is five-class macro-F1.
We train for 40 epochs with AdamW (lr\,=\,$10^{-4}$, weight decay $5{\times}10^{-4}$), batch size 128, cosine scheduling with 3-epoch warmup, and early stopping on validation action F1 (patience 15).
IMU signals are globally z-score normalized using training-set statistics and per-video centered to reduce device drift.
During training, we apply random augmentation: jittering ($\sigma{=}0.02$), uniform scaling (0.9--1.1), temporal masking, and small 3D rotation (${\pm}15^{\circ}$).
We compare against four baseline architectures that span the design space explored in recent head-mounted IMU work~\cite{egocharm2025}.
IMU2CLIP~\cite{imu2clip2023} uses the stacked CNN-GRU encoder designed for CLIP-aligned IMU pretraining.
CNN-LSTM-GRU uses LSTM-GRU sequence modeling without Transformer aggregation.
CNN-MLP replaces the temporal encoder and aggregator with MLPs.
MLP-MLP uses multi-layer perceptrons throughout.
All baselines are trained on the same data with identical focal loss and class weights.
We also evaluate Mantis~\cite{mantis2025}, a general time-series foundation model (8M params), with frozen features + SVM and adapter-head fine-tuning on action classification only (no scenario head).

\subsection{Main Results}
\label{sec:experiments:main}

\begin{table}[t]
\centering
\caption{Main results on 5-class behavioral-level action recognition. F1 is macro-averaged; Acc is micro-averaged. Mantis has no scenario head (---).}
\label{tab:main}
\resizebox{\columnwidth}{!}{%
\begin{tabular}{@{}lccccr@{}}
\toprule
\textbf{Model} & \textbf{Act.\ F1} & \textbf{Act.\ Acc} & \textbf{Scen.\ F1} & \textbf{Scen.\ Acc} & \textbf{Params} \\
\midrule
MLP-MLP                              & 0.339 & 0.436 & 0.528 & 0.578 & 1.03M \\
Mantis-8M frozen~\cite{mantis2025}   & 0.301 & 0.321 & ---   & ---   & 8M \\
IMU2CLIP~\cite{imu2clip2023}         & 0.385 & 0.435 & 0.559 & 0.590 & 4.0M \\
Mantis-8M fine-tuned                 & 0.370 & 0.467 & ---   & ---   & 8M \\
CNN-MLP                              & 0.367 & 0.451 & 0.523 & 0.554 & 1.07M \\
CNN-LSTM-GRU                         & 0.378 & 0.455 & 0.508 & 0.551 & 1.19M \\
\midrule
\textbf{HiT-HAR (ours)}             & \textbf{0.457} & \textbf{0.490} & \textbf{0.569} & \textbf{0.601} & \textbf{703K} \\
\bottomrule
\end{tabular}%
}
\end{table}

Table~\ref{tab:main} shows that HiT-HAR achieves the highest action F1 (0.457) and action accuracy (0.490) among all models, outperforming CNN-LSTM-GRU by 0.069 F1 and IMU2CLIP by 0.094 F1 while using 5.7$\times$ fewer parameters than IMU2CLIP.
Mantis-8M, despite having 11$\times$ more parameters, reaches only 0.370 action F1 with adapter fine-tuning, suggesting that Mantis's general-purpose time-series pre-training transfers poorly to behavioral-level IMU recognition.
The gap between macro-F1 and accuracy reflects class imbalance: accuracy is dominated by frequent classes, while F1 gives equal weight to each class including the rare Search category.
Per-class analysis reveals a clear observability hierarchy: Locomotion (F1\,=\,0.596) is reliably detected through gait periodicity, Object Transfer (0.519) and Task Operation (0.510) are partially separable with temporal context, Stationary (0.386) is moderate, and and Search (0.273), the most application-relevant class for proactive AR, marks the boundary where complementary sensors such as eye tracking would yield the greatest benefit.
This per-class hierarchy directly mirrors the raw-IMU separability analysis in Sec.~\ref{sec:observability}.

\subsection{Ablation Study}
\label{sec:experiments:ablation}

\paragraph{Multi-task weighting ($\beta$ sweep).}

We select $\beta{=}0.3$ as the balanced operating point: action F1 (0.457) is only 0.009 below the action-only optimum ($\beta{=}0$), while scenario F1 improves from 0.121 to 0.569.
For AR assistance, scenario context has practical value in enabling context-aware guidance.

\paragraph{Architecture scaling.}
Expanding HiT-HAR to 1.1M parameters (3-layer WAT, wider hidden dimensions) yields no improvement in action F1 (0.457 in both configurations).
Language-based label alignment (SBERT, CLIP, Qwen text encoders) also provides negligible gains (+0.004 at best), in line with Haresamudram~\etal~\cite{haresamudram2025limits}.
These results confirm that exploiting temporal context and scenario structure matters more than scaling model size.

%% file: sec/6_limits.tex
\section{Observability Frontier Analysis}
\label{sec:observability}

Beyond model performance, we ask: \emph{what can head-mounted IMU physically distinguish?}
We map the observability frontier through two complementary analyses, characterizing the boundaries of what head-mounted sensing can resolve and where complementary modalities would be needed.

\subsection{Per-Class-Pair IMU Separability}
\label{sec:observability:mmd}

\begin{figure}[!ht]
    \centering
    \includegraphics[width=\linewidth]{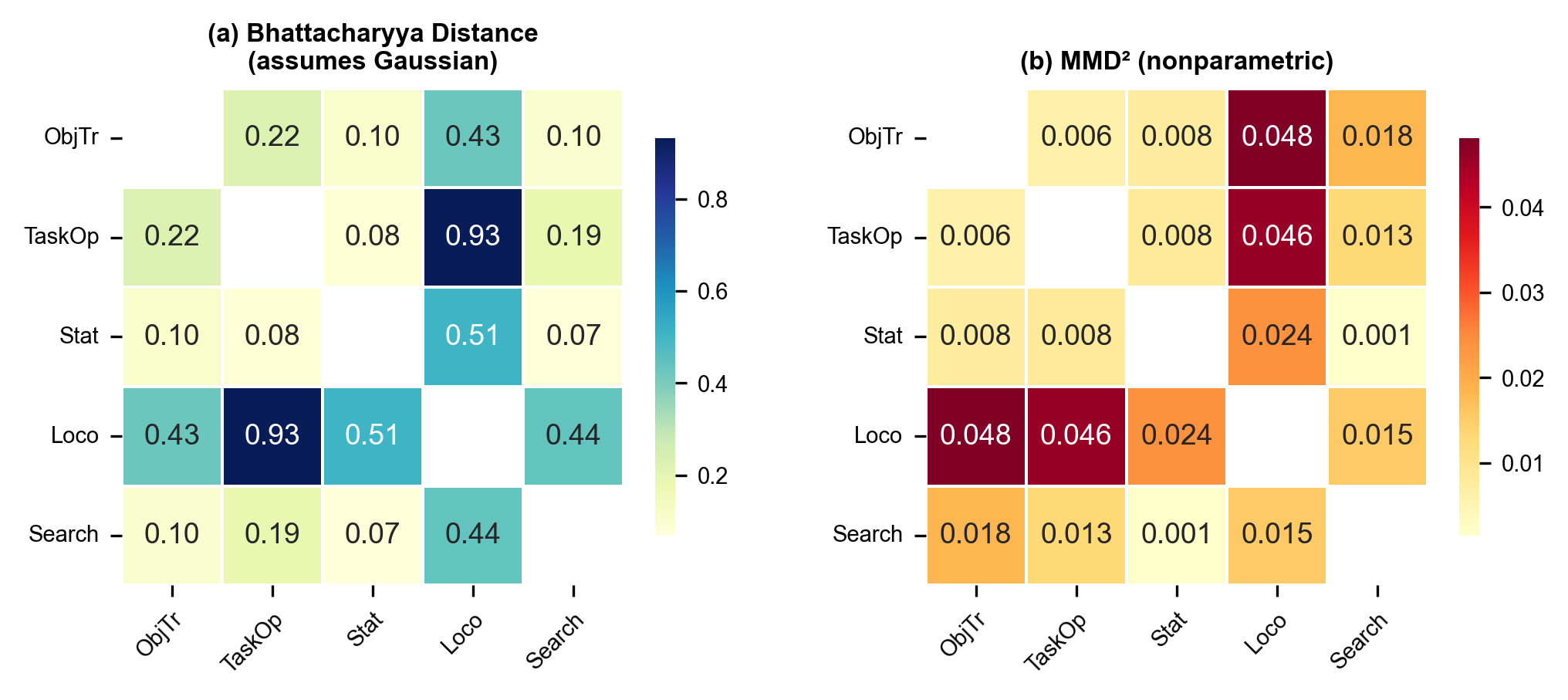}
    \caption{Pairwise class separability in 8-dim IMU feature space. (a) Bhattacharyya distance: Locomotion stands out (3.3--5.0). (b) MMD$^2$ ($n{=}1{,}000$ permutations, Bonferroni-corrected): five pairs are significant ($p < 0.05$).}
    \label{fig:mmd}
\end{figure}

We measure pairwise class separability directly in the 8-dimensional 
raw IMU feature space (6-axis plus 2 norms) using two complementary 
metrics (Fig.~\ref{fig:mmd}).
The \emph{Bhattacharyya distance} quantifies distributional overlap 
under a Gaussian assumption (higher values indicate better 
separability), while the nonparametric \emph{MMD two-sample test} 
assesses whether two class distributions are statistically 
distinguishable without distributional assumptions (permutation-based 
$p$-values, $n{=}1{,}000$, Bonferroni-corrected).

Both metrics reveal the same structure.
Locomotion separates cleanly from all four other classes 
(Bhattacharyya 3.3--5.0, $\text{MMD}^2$ 0.016--0.048, all 
$p < 0.05$), and Object Transfer separates from Search 
($p = 0.020$), yielding 5 of 10 class pairs
statistically distinguishable from head motion alone.
The hardest pairs remain Object Transfer vs.\ Task Operation 
($\text{MMD}^2 = 0.006$, $p = 0.94$) and Stationary vs.\ Search 
($\text{MMD}^2 = 0.001$, $p = 1.00$), confirming that these class pairs have substantial overlap in per-window IMU features and that temporal aggregation, as in HiT-HAR, is needed to exploit sequential context.

However, separability varies by scenario: in scenarios with frequent object displacement (e.g., Cooking, Carpentry), the head-turning pattern during Object Transfer becomes more distinctive, while Task Operation involves a steadier gaze. Similarly, Search and Stationary overlap because most Search instances involve static visual scanning, but Search during outdoor walking produces more head rotation and is better separable. These scenario-dependent patterns motivate the multi-task design of HiT-HAR, where scenario context aids per-window action 
classification.

\begin{figure*}[!t]
    \centering
    \begin{minipage}{0.48\textwidth}
        \centering
        \includegraphics[width=\linewidth]{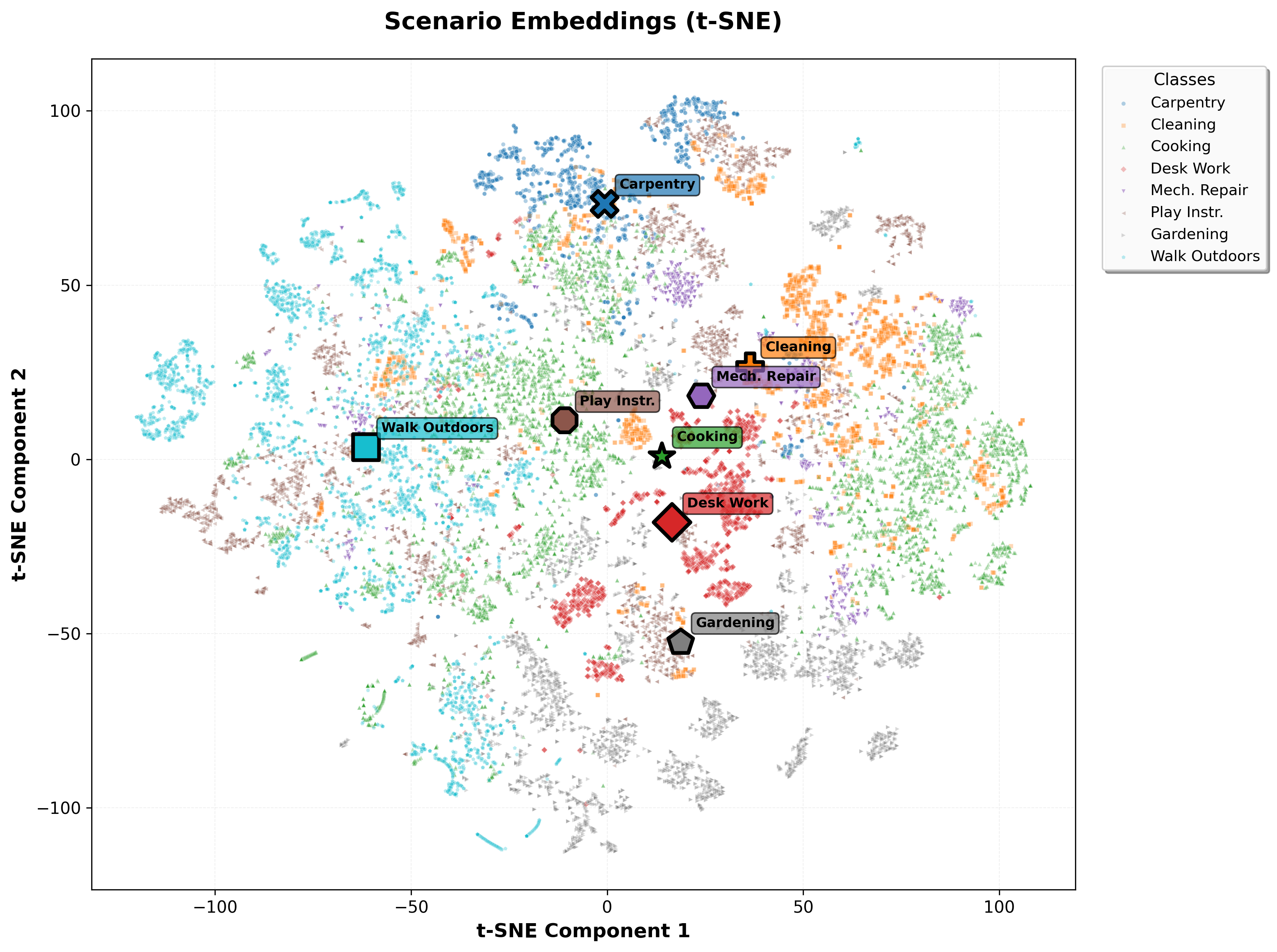}
        \centerline{\small (a) Scenario embeddings}
    \end{minipage}
    \hfill
    \begin{minipage}{0.48\textwidth}
        \centering
        \includegraphics[width=\linewidth]{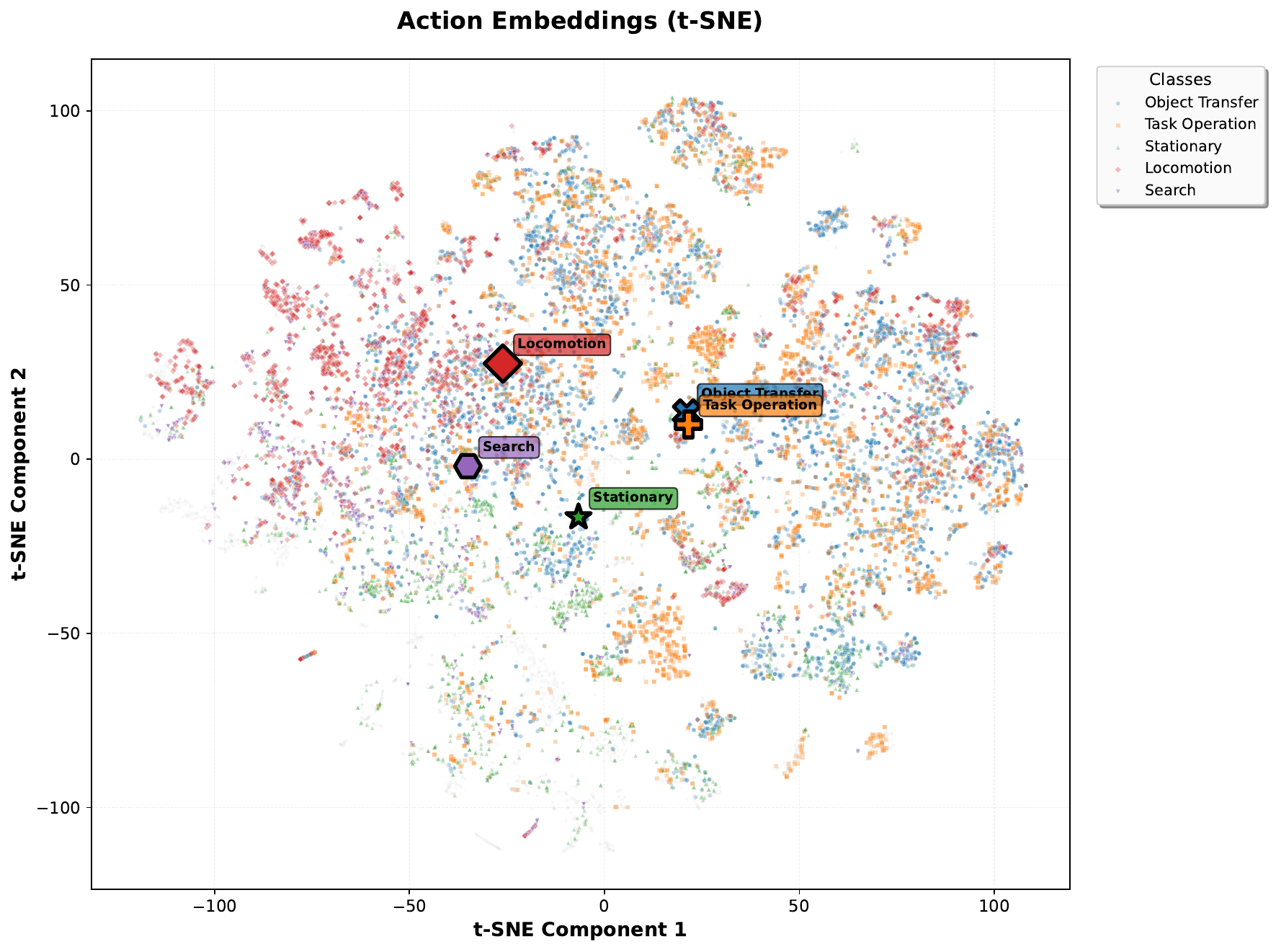}
        \centerline{\small (b) Action embeddings}
    \end{minipage}
    \caption{t-SNE of HiT-HAR learned embeddings ($\beta{=}0.3$). (a) Scenario embeddings form distinct clusters. (b) Action embeddings show regions of class overlap.}
    \label{fig:tsne}
\end{figure*}

\subsection{Learned Embedding Structure}
\label{sec:observability:tsne}

Fig.~\ref{fig:tsne} visualizes the learned embeddings via t-SNE.
Scenario embeddings form recognizable clusters (Walk Outdoors, Carpentry, Desk Work are well separated), explaining the strong scenario F1 of 0.569.
Action embeddings are far more entangled: Object Transfer, Task Operation, Stationary, and Search overlap extensively, with only Locomotion showing partial separation, corroborating the MMD analysis above.

\begin{figure}[t]
    \centering
    \includegraphics[width=\linewidth]{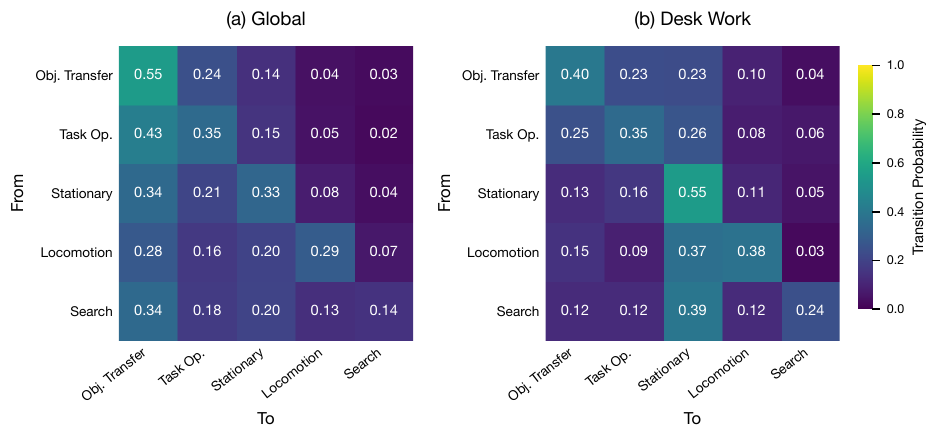}
    \caption{Behavioral state transition probabilities. (a) Global: Object Transfer and Task Operation form a dominant cycle. (b) Desk Work: Stationary dominates, reflecting a structurally different behavioral dynamic.}
    \label{fig:transitions}
\end{figure}

\subsection{Temporal Transition Structure}
\label{sec:observability:transitions}

Beyond per-window separability, behavioral states form structured transition sequences that vary by scenario.
Fig.~\ref{fig:transitions} shows the global transition matrix alongside Desk Work as a contrasting scenario.
Object Transfer and Task Operation form a dominant off-diagonal cycle (0.24/0.43), reflecting the natural fetch-use-fetch workflow in manipulation-heavy scenarios.
Desk Work shows a fundamentally different pattern: Stationary self-transition jumps from 0.33 to 0.55, and Locomotion transitions primarily to Stationary (0.37) rather than to manipulation classes.
These scenario-dependent temporal patterns further justify multi-task scenario supervision in HiT-HAR.

\subsection{Summary}
\label{sec:observability:summary}

The separability analysis, embedding structure, and transition patterns collectively delineate an observability frontier: Locomotion is fully observable through gait periodicity, Object Transfer and Task Operation become separable with temporal context and scenario supervision, and Search shows the strongest scenario dependence with better separability in outdoor active-scanning scenarios.

%% file: sec/8_conclusion.tex
\section{Discussion and Conclusion}
\label{sec:conclusion}

We push head-mounted IMU beyond motion primitives by defining five behavioral categories that balance application need with sensor capability, then systematically mapping the observability boundaries of this modality.
HiT-HAR, a 703K-parameter hierarchical model, outperforms established IMU-HAR baselines on action recognition (Table~\ref{tab:main}), while ablations confirm that model scaling provides negligible returns.

The observability frontier analysis confirms a clear hierarchy from reliably observable (Locomotion) through context-dependent (Object Transfer, Task Operation), with scenario-dependent signal overlap as the remaining challenge for classes like Search.
Architectural choices that exploit temporal context and scenario structure prove more effective than simply scaling model capacity for behavioral-level recognition from head-mounted IMU.

\paragraph{Limitations.}
We use a single head-mounted IMU; wrist or body sensors could provide complementary signals.
Action labels cover 17.4\% of total video time; the remaining unlabeled time could support self-supervised pretraining.
Our labels are derived from narration timestamps that mark when an action is mentioned rather than its full duration, so a quick manipulation and a sustained one both occupy the same 1-second window.
We evaluate offline on recorded data; real-time inference latency and on-device deployment remain to be validated.

\paragraph{Future work.}
The observability gaps identified here point toward targeted sensor fusion — eye tracking to resolve Search from Stationary, wrist IMU for separating manipulation classes, and audio for disambiguating Task Operation subtypes.
Self-supervised pretraining on the unlabeled video time, alongside systematic exploration of temporal window lengths and alternative architectures, can probe the current frontier further.
The structured transition patterns in Sec.~\ref{sec:observability:transitions} also suggest that lightweight next-state forecasting models could enable proactive AR assistance, leveraging the non-uniform, scenario-dependent transitions as a training signal.
Because our annotations are paired with Ego4D video, the same behavioral taxonomy can serve as supervision for cross-modal tasks.

%% file: sec/X_suppl.tex
\clearpage
\appendix
\twocolumn[{\centering\Large\textbf{Appendix}\vspace{1em}\par}]

\section{Additional IMU Signal Visualizations}
\label{sec:suppl:imu}

\begin{figure}[h!]
    \centering
    \includegraphics[width=\linewidth]{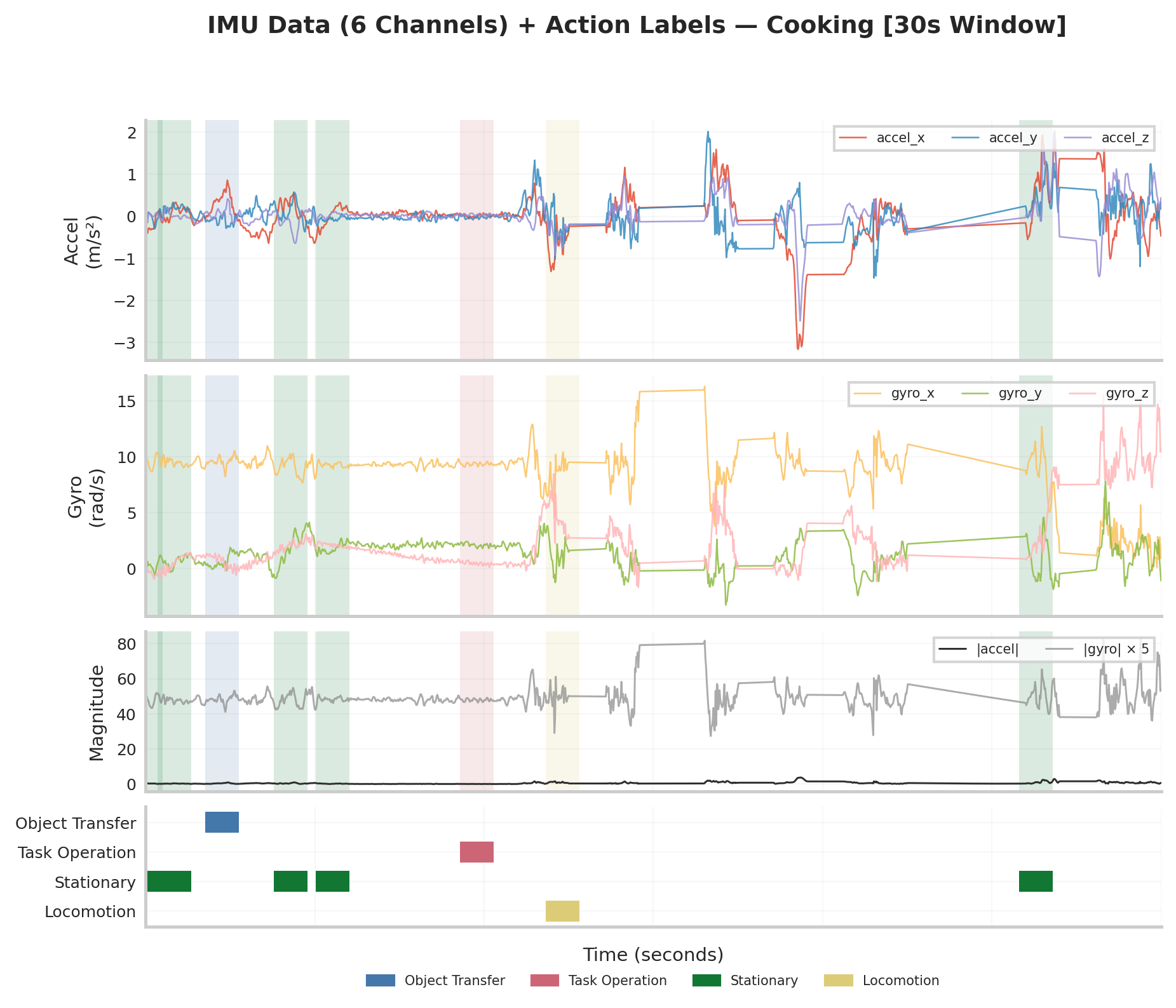}
    \caption{Head-mounted IMU signals (30-second window, Cooking) with 5-class action labels. Frequent Object Transfer and Stationary transitions reflect the pick-prepare-pause workflow.}
    \label{fig:imu_cooking}
\end{figure}

\begin{figure}[h!]
    \centering
    \includegraphics[width=\linewidth]{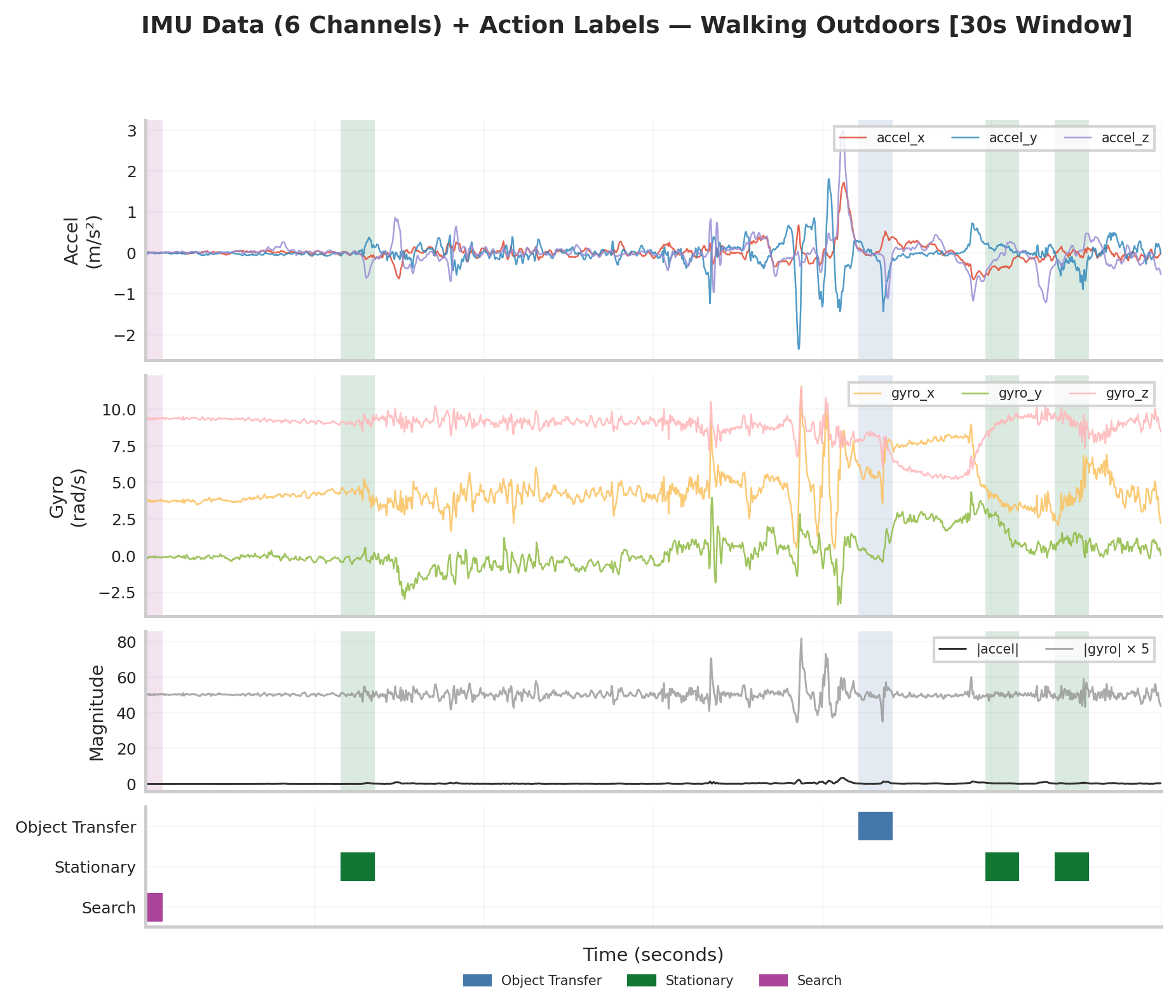}
    \caption{Head-mounted IMU signals (30-second window, Walking Outdoors) with 5-class action labels. Locomotion produces distinctive periodic gait patterns in the accelerometer channels.}
    \label{fig:imu_walking}
\end{figure}

Fig.~\ref{fig:imu_cooking} and Fig.~\ref{fig:imu_walking} show 30-second head-mounted IMU windows from two additional Ego4D scenarios, complementing the Mechanical Repair example in the main paper (Fig.~\ref{fig:imu_example}).

\section{Data Quality Analysis}
\label{sec:suppl:quality}

\begin{figure}[h!]
    \centering
    \includegraphics[width=\linewidth]{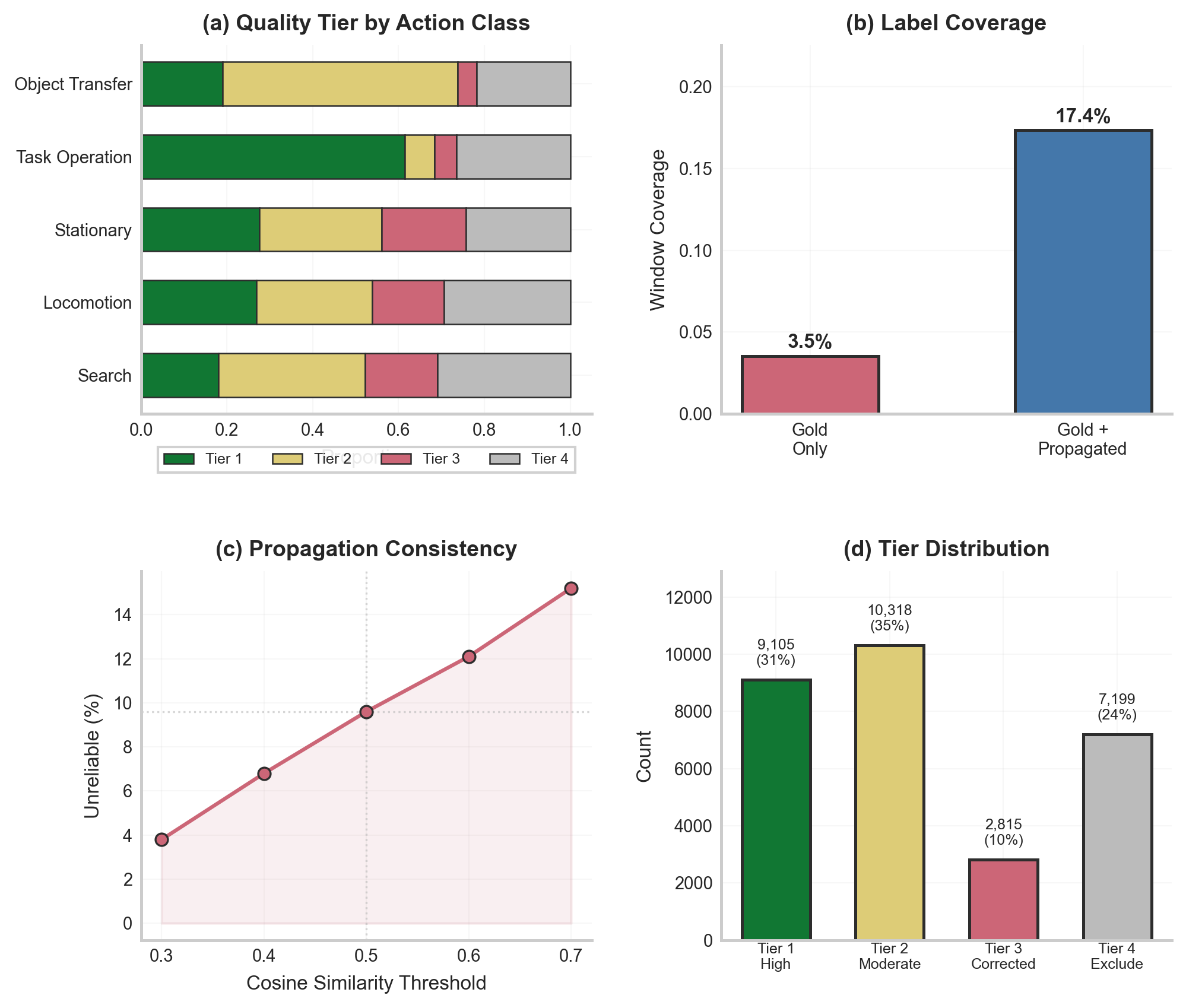}
    \caption{Annotation quality analysis. (a) Quality tier distribution by action class. (b) Label coverage: gold-only (3.5\%) vs.\ gold + propagated (17.4\%). (c) Propagation consistency vs.\ cosine similarity threshold. (d) Overall tier distribution (27K gold annotations).}
    \label{fig:data_quality}
\end{figure}

\begin{figure}[h!]
    \centering
    \includegraphics[width=\linewidth]{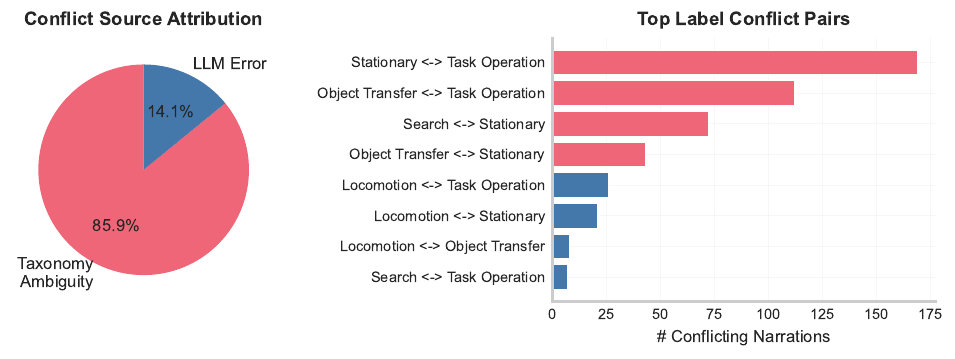}
    \caption{Label conflict source attribution. Left: 85.9\% of multi-label conflicts arise from taxonomy boundary ambiguity, not LLM errors. Right: top conflict pairs ranked by frequency.}
    \label{fig:conflict_source}
\end{figure}

Fig.~\ref{fig:data_quality} details the annotation quality framework from Sec.~\ref{sec:dataset:quality}, showing per-class tier distributions, label coverage before and after propagation, and propagation consistency.
Fig.~\ref{fig:conflict_source} breaks down the 470 multi-label conflicts by source, confirming that the dominant challenge is taxonomy boundary ambiguity rather than LLM errors.

\section{Per-Window Feature Ceiling}
\label{sec:suppl:ceiling}

\begin{figure}[h!]
    \centering
    \includegraphics[width=0.85\linewidth]{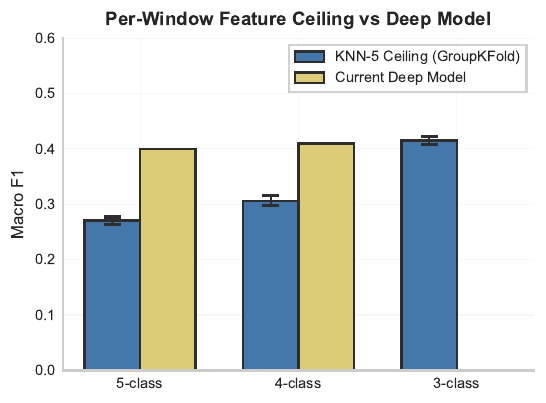}
    \caption{Per-window feature ceiling (KNN-5, GroupKFold) vs.\ deep model macro F1 across 5-class, 4-class, and 3-class taxonomies. The deep model exceeds the per-window ceiling, demonstrating the value of temporal aggregation.}
    \label{fig:ceiling}
\end{figure}

Fig.~\ref{fig:ceiling} compares the per-window feature ceiling (KNN-5 with GroupKFold on 42 statistical IMU features) against HiT-HAR across three taxonomy granularities.
The deep model exceeds the per-window ceiling in all configurations, confirming that temporal aggregation captures patterns beyond what single-window statistical features can represent.

\section{Task-Weighting Sensitivity Analysis}
\label{sec:suppl:beta}

\begin{figure}[t]
    \centering
    \includegraphics[width=0.85\linewidth]{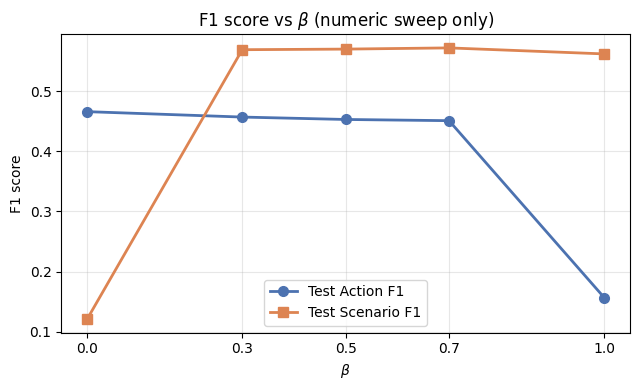}
    \caption{Test macro F1 vs.\ task-weighting coefficient $\beta$. Scenario F1 saturates at $\beta{\geq}0.3$; action F1 remains stable until $\beta{=}1.0$, where loss is entirely scenario-driven and action performance collapses.}
    \label{fig:beta_sweep}
\end{figure}

Fig.~\ref{fig:beta_sweep} shows the full $\beta$ sweep across five values.
At $\beta{=}0$, the model is action-only and scenario F1 is near chance (0.12); introducing even moderate scenario supervision ($\beta{=}0.3$) raises scenario F1 to 0.57 while preserving action performance.
Action F1 remains stable for $\beta \in [0.0, 0.7]$ but collapses at $\beta{=}1.0$, where the loss is scenario-dominated and the action head receives no direct gradient signal.
This confirms that the optimal operating point lies at low $\beta$ values where the scenario auxiliary task regularizes without degrading the primary action objective.

\begin{figure}[t]
    \centering
    \includegraphics[width=0.85\linewidth]{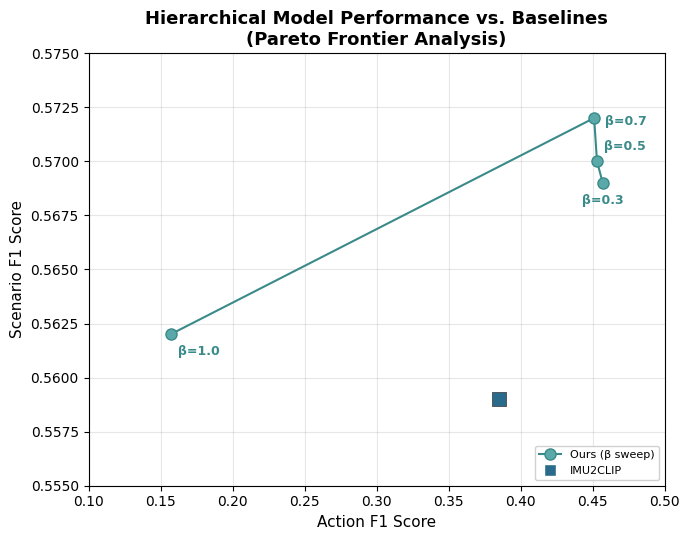}
    \caption{Pareto frontier in the action--scenario F1 plane. HiT-HAR ($\beta \in \{0.3, 0.5, 0.7\}$) strictly dominates IMU2CLIP on both tasks simultaneously.}
    \label{fig:pareto}
\end{figure}

Fig.~\ref{fig:pareto} reframes the same sweep as a Pareto frontier in the action--scenario F1 plane.
HiT-HAR at $\beta \in \{0.3, 0.5, 0.7\}$ Pareto-dominates IMU2CLIP on both axes, demonstrating that hierarchical multi-task learning achieves a strictly better trade-off than the strongest single-task baseline.

\section{Model Efficiency}
\label{sec:suppl:efficiency}

\begin{figure}[t]
    \centering
    \includegraphics[width=0.85\linewidth]{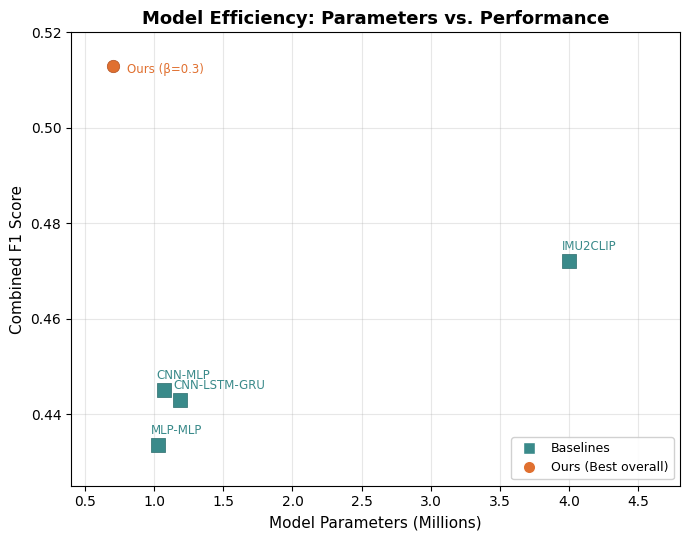}
    \caption{Model efficiency: combined F1 vs.\ parameter count (millions). HiT-HAR achieves the best performance at the smallest model size.}
    \label{fig:efficiency}
\end{figure}

Fig.~\ref{fig:efficiency} plots combined F1 (mean of action and scenario macro F1) against model size for all evaluated architectures.
HiT-HAR ($\beta{=}0.3$) achieves the highest combined F1 (0.51), while IMU2CLIP requires $4{\times}$ more parameters to reach a lower combined score (0.47).
The lightweight baselines (MLP-MLP, CNN-MLP, CNN-LSTM-GRU) cluster at ${\sim}1$M parameters but trail by 6--8 points in combined F1, confirming that HiT-HAR's gains stem from architectural design rather than parameter scaling.

\section{Complete Hyperparameter Configuration}
\label{sec:suppl:hyperparams}

Table~\ref{tab:hyperparams} lists all hyperparameters used for training HiT-HAR (headline model: \texttt{beta\_sweep\_5class\_b03}).

\begin{table}[h!]
\centering
\caption{Complete hyperparameter configuration for HiT-HAR.}
\label{tab:hyperparams}
\small
\setlength{\tabcolsep}{4pt}
\begin{tabular}{@{}llr@{}}
\toprule
\textbf{Component} & \textbf{Parameter} & \textbf{Value} \\
\midrule
\multicolumn{3}{@{}l}{\emph{Window-Level Encoder (WLE)}} \\
& Input channels & 8 \\
& Embedding dim & 128 \\
& Window size (samples) & 50 \\
& CNN dilations & \{1, 2, 4\} \\
& SE reduction ratio & 8 \\
& BiGRU hidden size & 96 \\
& Attention pooling dim & 64 \\
& Dropout & 0.3 \\
\midrule
\multicolumn{3}{@{}l}{\emph{Window Aggregation Transformer (WAT)}} \\
& Transformer layers & 1 \\
& Attention heads & 4 \\
& Feed-forward dim & 512 \\
& Positional embeddings & Learnable \\
& Activation & GELU \\
& Norm & Pre-LN \\
& Sequence length & 30 \\
& Dropout & 0.2 \\
\midrule
\multicolumn{3}{@{}l}{\emph{Training}} \\
& Optimizer & AdamW \\
& Learning rate & $10^{-4}$ \\
& Weight decay & $5 \times 10^{-4}$ \\
& Batch size & 128 \\
& Epochs (max) & 40 \\
& Warmup epochs & 3 \\
& Scheduler & Cosine (min factor 0.2) \\
& Gradient clipping & max norm 1.0 \\
& EMA decay & 0.999 \\
& Label smoothing $\epsilon$ & 0.05 \\
& Early stopping (patience) & 15 \\
\midrule
\multicolumn{3}{@{}l}{\emph{Loss}} \\
& Task weighting $\beta$ & 0.3 \\
& Focal loss $\gamma$ & 2.0 \\
& Action class weights & [0.95, 1.0, 1.6, 1.2, 3.0] \\
& Scenario class weights & [1.9, 1.0, 1.4, 2.0, 7.6, \\
&                         & 1.2, 2.1, 1.6] \\
\midrule
\multicolumn{3}{@{}l}{\emph{Data}} \\
& Normalization & Global z-score \\
& Per-video centering & Yes \\
& Augmentation: jitter $\sigma$ & 0.02 \\
& Augmentation: scaling & [0.9, 1.1] \\
& Augmentation: rotation & $\pm 15^{\circ}$ \\
& Window stride & 10 \\
\bottomrule
\end{tabular}
\end{table}